\documentclass{Interspeech2024}
\usepackage{hyperref}
\usepackage{xurl}
\usepackage{breakurl}
\usepackage{tabularx}
\usepackage{stfloats}
\usepackage{algorithm} 
\usepackage{algpseudocode} 
\usepackage{adjustbox}
\usepackage{svg}
\svgpath{{./img/}} 

\interspeechcameraready

\newcommand{\mname}{FASA}

\title{FASA: a Flexible and Automatic Speech Aligner for Extracting High-quality Aligned Children Speech Data}

\name[affiliation={1}]{Dancheng}{Liu}
\name[affiliation={1}]{Jinjun}{Xiong}

\address{
  $^1$SUNY Buffalo, USA}
\email{dliu37@buffalo.edu, jinjun@buffalo.edu}

\keywords{automatic speech recognition, forced-alignment, children ASR, dataset, deep learning}
\begin{document}

\maketitle
 
\begin{abstract}
Automatic Speech Recognition (ASR) for adults' speeches has made significant progress by employing deep neural network (DNN) models recently, but improvement in children's speech is still unsatisfactory due to children's speech's distinct characteristics. 
DNN models pre-trained on adult data often struggle in generalizing to children's speeches with fine tuning because of
the lack of high-quality aligned children's speeches.
When generating datasets, human annotations are not scalable, and existing forced-alignment tools are not usable as
they make impractical assumptions about the quality of the
input transcriptions. To address these challenges,
we propose a new forced-alignment tool, FASA, as
a flexible and automatic speech aligner to extract
high-quality aligned children's speech data from
many of the existing noisy children's speech data. We demonstrate
its usage on the CHILDES dataset and show that FASA can improve
data quality by 13.6$\times$ over human annotations. 
The toolkit is available at: \url{https://github.com/DanchengLiu/FASA}.
\end{abstract}

\section{Introduction}
In recent years, the field of Automatic Speech Recognition (ASR) has witnessed remarkable advancement in light of deep learning (DL), transforming our interactions with technology and enhancing various applications, from virtual assistants \cite{BERNARD2019VAPhil,Brill2019VAsurvey,Christophe2023ASR_re} to transcription services \cite{Joshua2019TSnon-verbal}. While these developments have significantly benefited adults, the application of ASR  to children's speech presents a unique set of challenges. Children's speech exhibits distinct characteristics, including rapid changes in pitch, articulation patterns, and vocabulary development \cite{Bhardwaj2022reviewASR}.
These differences make fine-tuning deep neural networks (DNNs) pre-trained on adult speeches hard to be 
generalized for children's speech.

Although previous studies \cite{jain2023adaptation,Wills2023non-native,Prashanth2022Eval,Shraddha2022Eval,booth-etal-2020-evaluating,rolland-etal-2022-multilingual} have demonstrated 
some success in fine-tuning pre-trained ASR models for children ASR, 
it requires large and diverse high-quality children
speech data, i.e., the transcription and the audio are
matched (or aligned) well. 
Given children speech's distinctive speech patterns, especially for children with speech and language disorders,
it is hard to obtain such high-quality data through
human efforts. 
On the one hand, human annotation for children's speech
is labor-intensive and would often require domain expertise. Both our own experience and previous work~\cite{Miller_Andriacchi_Nockerts_2016} showed
that annotation for an audio segment
would take 3 to 8 times more than the original audio's duration.
Moreover, human annotation is often task-specific and varies significantly in its quality. For example,
there are many existing transcribed children's speech data, such as those from CHILDES~\cite{CHILDES} for children with speech and language disabilities, but their transcription purposes are quite different, with some focusing on stuttering,
some on speech sound disorder, and some on dialects.
This made those transcriptions full of additional
task-specific annotations, and sometimes the transcriptions are even incomplete because of the lack of relevance to the specific tasks.


A natural question arising is whether we could develop a tool that can extract high-quality children's speech datasets from many of the existing low-quality children's speech datasets and a natural answer is forced alignment.
forced alignment is the process of aligning audio segments with their transcriptions. However, previous forced
alignment methods like MFA \cite{mcauliffe17_interspeech} rely on the correctness of the provided transcriptions, 
which is often impractical to obtain for many datasets. 
To address this issue and facilitate further research in children's ASR, we propose to develop a new forced-alignment tool
that is both flexible and automated so that it can
consider a wide range of issues in existing children
speech datasets.
The major contributions of this paper are as follows.
First, we present a novel open-sourced forced-alignment toolkit, Flexible and Automatic Speech Aligner (\mname), to
extract accurate, aligned, well-segmented audio segments and the corresponding transcriptions under flexible conditions. 
Second, we leverage state-of-the-art DL models as the backbone to help with the alignment under minimal requirements on the provided transcriptions. 
Third, we use FASA to compile many new and high-quality children's speech datasets based on CHILDES~\cite{CHILDES}. To the best of our knowledge, this is the first large-scale well-aligned children's speech datasets, including clinical data. Our experiments show the superiority and consistency of \mname~ when compared to existing forced-alignment toolkits and even human annotators.

The rest of the paper is organized as follows. Section 2  formally defines the forced-alignment problem, and introduces the related works in the area of ASR and alignment tools. Section 3 describes \mname ~in detail. We show in Section 4  the superior performance of this new toolkit under noisy and unorganized datasets such as CHILDES~\cite{CHILDES}.
Finally, we conclude the paper in Section 5. The code and instructions will be open-sourced after publication.

\section{Related Works}

\subsection{Definition of the Task}
Given a non-timestamped audio and its non-timestamped transcription (that may be noisy and incomplete), forced alignment will produce a set of time-stamped audio segments with their corresponding time-stamped high-quality transcriptions. Modern ASR systems expect the input audio to be segmented into smaller pieces during their training process. For example, the Whisper model \cite{radford2022whisper} pads or trims the audio input to 30 seconds. Thus, when a transcription without a timestamp is associated with a long audio, a forced-alignment toolkit is essential for the creation of such a dataset. Formally, the forced-alignment task is defined over an audio sample composed of $n$ utterances, $A:=\{A_1,A_2,...A_n\}$, and a transcription of $m$ ``words", $T:=\{T_1,T_2,...T_m\}$. It is noteworthy that a ``word" in $T$ refers to the basic element of the transcription. Depending on the use case, it could represent a sentence, a word, or even a phonetic symbol. The forced-alignment task is to associate each $A_i$ with its corresponding words in $T$, which are from $T_{si}$ to $T_{ei}$, or report that $A_i$ is not transcribed in $T$. For ease of notation, we will define the association between $A_i$ and its transcription as $A_i = (T_{si}, T_{ei})$. 

A robust auto-alignment system will need to have two important features. First, it shall not assume that if $A_i= (T_{si}, T_{ei})$, $A_j= (T_{sj}, T_{ej})$, and $i<j$, then $ei<sj$. That is, the transcription does not have time-dependency with the audio. An utterance appearing early in the audio does not necessarily mean it appears early in the transcription. Second, $A_i$ does not necessarily have a corresponding $(T_{si}, T_{ei})$, and it is possible that $A_i=\emptyset$. That is, not all audio information will be transcribed into the transcription, and some audio will be left un-transcribed. Similarly, $T_k\in T$ does not imply $T_k\in A$. That is, not all words in the provided transcription have corresponding audio.

Such two features are important because they do not require the completeness and orderliness of the provided transcription, which is usually a more practical scenario in the real-world. A popular approach for obtaining large amounts of paired audio and transcriptions is by scraping the Internet, but the vast majority of transcriptions from the Internet would be noisy and incomplete, containing missing and 
sometimes wrongly-ordered transcriptions. Under those circumstances, existing forced-alignment toolkits such as MFA \cite{mcauliffe17_interspeech} fall short, and we will discuss this in more detail in Section 2.2.

\subsection{Existing Alignment Tools}
Traditionally yet still prevalently, alignment between the audio and its transcription is done via human annotators on various software \cite{praat,grover2020audino}. However, as discussed earlier, such practice is not scalable for large datasets. Work from \cite{KISLER2017326} contains some parts of a complete forced-alignment pipeline, but it does not address the fundamental problem of aligning audio with its transcription.
Work from \cite{GAN2019} uses generative-adversarial networks (GAN) to perform data augmentation on children ASR dataset, but their work does not introduce diverse \textit{new} data to the field. Recently, the Talkbank project announced its data processing pipeline that converts raw audio into CLAN-annotated transcriptions \cite{Liu_MacWhinney_Fromm_Lanzi_2023}. While their work uses a similar backbone structure as ours, they rely on transcription generated by ASR models, whereas we faithfully adhere to the provided transcription as the ground truth. Thus, on downstream tasks such as fine-tuning ASR models, our dataset will be more usable because datasets generated by ASR models might cause severe degradation according to \cite{radford2022whisper}. On the other hand, there have been several works on forced-alignment ASR datasets with the assistance of human-labeled transcriptions \cite{mcauliffe17_interspeech, POnSS, zhang2023speechgpt}, with Montreal-Forced-Aligner (MFA) being the most popular toolkit \cite{mcauliffe17_interspeech}. MFA incorporates Kaldi \cite{Povey_ASRU2011} as the backbone, which uses the Gaussian Mixture Model (GMM) for its transcription generation process. However, while MFA \cite{mcauliffe17_interspeech} works well with carefully annotated transcriptions, it requires the transcription to have a perfect matching with the audio. That is, $A_1 \to \{T_1,T_2,...,T_i\}$, $A_2\to \{T_{i+1},T_{i+2},...,T_{j}\}$, and so forth. 
Our experiments in Section 4.2 show that MFA \cite{mcauliffe17_interspeech} will not be functional on noisy transcriptions as discussed in Section 2.1. Moreover, while recent multi-modal large language models (MLLM) might have the potential of automating the alignment process \cite{zhang2023speechgpt}, they are much more resource-intensive compared to specific ASR models.
\subsection{Dataset}

Recent adaptions towards children ASR usually consider general datasets such as PF-STAR \cite{batliner05b_interspeech}, My Science Tutor (MyST) \cite{pradhan2023science}, CMU Kids Corpus \cite{CMU_kids}, OGI Kids Corpus \cite{OGI}, or smaller-scale clinical datasets for children with older ages such as ETLT, a German-English ASR dataset for older children \cite{gretter21_interspeech}. Among them, MyST is the largest dataset, containing 393 hours of speech data between children and a virtual science tutor, which is still significantly smaller than the adult datasets. We've observed that the majority of children's ASR models are typically fine-tuned using general datasets, leading to a lack of task-specific abilities in crucial areas, such as clinical settings for young children. This limitation arises from the unavailability of high-quality datasets tailored to these specific task requirements. However, we are optimistic that our toolkit will address and resolve this issue effectively.

\section{\mname~Toolkit Design}


\newcommand{\Input}[1]{\algrenewcommand{\alglinenumber}[1]{Input: \ \setcounter{ALG@line}{\numexpr##1-1}} #1}
\newcommand{\Step}[1]{\algrenewcommand{\alglinenumber}[1]{Step ##1: } #1}
\newcommand{\NoNumber}{\algrenewcommand{\alglinenumber}[1]{\setcounter{ALG@line}{\numexpr##1-1} \ \ \ \ \ \ \ \ \ \ }}
\newcommand{\Output}[1]{\algrenewcommand{\alglinenumber}[1]{Output:\setcounter{ALG@line}{\numexpr##1-1}} #1}
\begin{algorithm*}[!ht]
	\caption{sliding window to find the best matching} 
    \label{algo:algo}
	\begin{algorithmic}[1]
        
         \Input \State $A$, $T=\{T_1,...T_m\}$, $\bar{T}$, alignment threshold $\sigma_a$, inclusion threshold $\sigma_i$.
            \Step \State Initialize holder for dataset of aligned segments: $\text{DATA}_{align}$ = []
            \NoNumber \State Initialize holder for questionable segments:  $\text{DATA}_{verify}$ = []
		\Step \For {$A_k \in A$}
                \NoNumber \State Get $A_k$'s transcription: $\bar{T_{A_k}} = \{\bar{T_{i}}...\bar{T_{j}}\}\ \in \bar{T}$
                \NoNumber \State Initialize minimum distance $D_{min} = \infty$, best starting index $BEST_i$, best length $BEST_l$
			\NoNumber\For {$a=1,2,\ldots,m$}
                        \For {$b=1,2,\ldots,(j-i)$}   
				\If {$DIS(\bar{T_{A_k}}, T[a:a+b+1]) < D_{min}$}
                        \State $D_{min}$ = $DIS(\bar{T_{A_k}}, T[a:a+b+1])$
                        \State $BEST_i$ = $a$
                        \State $BEST_l$ = $b+1$

                    \EndIf
			\EndFor
            \EndFor
            
            \Step \State let $GT_k = T[BEST_i:BEST_i+BEST_l]$
            \NoNumber\If {$WER$ ($GT_k$, $\bar{T_{A_k}}$) $< \sigma_i$}
                \NoNumber\If {$WER$ ($GT_k$, $\bar{T_{A_k}}$) $< \sigma_a$}
                    \NoNumber \State append $(A_k, GT_k)$ to $\text{DATA}_{align}$
                    \Else
                    \NoNumber \State append $(A_k, GT_k, \bar{T_{A_k}} )$ to $\text{DATA}_{verify}$
                \EndIf
            \EndIf
		\NoNumber \EndFor

    \Output \State $\text{DATA}_{align}$, $\text{DATA}_{verify}$
	\end{algorithmic} 
\end{algorithm*}

\subsection{Features of ~\mname}
Similar to existing auto-alignment toolkits, \mname~ requires an audio file and a transcription associated with the audio file. However, due to the high uncertainty in the raw dataset, \mname ~assumes only the minimum format from the input. In particular, \mname~ does not assume the correctness of the transcription. The ground truth (GT) of an utterance $A_k$ is defined by Equation \ref{eq} in the pipeline of \mname. Compared to previous forced-alignment toolkits, \mname~ will choose to ignore the utterances without proper transcriptions, and thus reach significantly better dataset quality from this practice.

\begin{equation}
GT_{k}=
\begin{cases}
  T_{A_k}=\{T_i,T_{i+1},...,T_j\}, & \text{if}\ T_{A_k} \in T \\
  \emptyset, & \text{otherwise}
\end{cases}
\label{eq}
\end{equation}

Besides the improvement in the flexibility of datasets, \mname~ incorporates beneficial design elements from established toolkits. In a bid to enhance user convenience, \mname~ follows the same design principles as MFA for its usage. Users simply have to prepare the audio file and its transcriptions into a designated folder, and then execute a program. The subsequent processes are all automatic, streamlining the user experience. \mname~ further integrates a crucial feature as present in both MFA \cite{mcauliffe17_interspeech} and PonSS \cite{POnSS}. This feature empowers users to select and manually input transcriptions for utterances in instances where the provided transcriptions raise suspicions of inaccuracy. This flexibility ensures precision and user control over the transcription process. At the same time, \mname~ incorporates an optional post-generation checking schema that enables automatic exclusion of incorrect alignments in the generated dataset, minimizing the possibility of incorrectness from the underlying model.

\begin{figure}[h]
  \centering
 \includegraphics[width=0.5\textwidth]{img/FASA_paper.png}
  \caption{This figure illustrates the pipeline of FASA. The input is an audio file and a transcription. Module \textcircled{1} optionally cleans the input transcription; module \textcircled{2} segments and makes predictions on the audio; module \textcircled{3} forced-aligns audio segments with the provided transcription using Algorithm 1; module \textcircled{4} performs post-generation checking (PGC); and module \textcircled{5} allows user to augment dataset via manual selections. 
  The entire system besides module \textcircled{5} is automatic.}
  \label{fig:pipe}
  \vspace{-7mm}
\end{figure}

\subsection{Workflow}
\mname~ follows a five-module pipeline to automatically segment, label, and align a long audio file with its transcription, as illustrated in Figure \ref{fig:pipe}. Among the five modules, the second and third are mandatory, whereas the other three are optional for enhancing the quality and quantity of the dataset. These five modules together maximize the correctness of forced alignment under flexible conditions. $\textbf{\textcircled{1}}$ The first module applies a regular expression to clean the provided transcriptions and to exclude any non-alphanumeric characters. $\textbf{\textcircled{2}}$ For the second module, modern ASR models will be used to obtain word-level timestamps of the transcriptions. 
Currently, sentence-level separations from the provided model are used as the segmentation marks for long audio. 
$\textbf{\textcircled{3}}$ After the second module, a folder consisting of audio segments and their corresponding predictions will be generated. The set of predictions for sentence-level utterances will be denoted as $\bar{T} = \{\bar{T_1}, \bar{T_2},...,\bar{T_n}\}$. For each utterance $A_k$, its predicted transcription will be $A_k=\bar{T_{A_k}}$. The third module will apply a sliding-window Algorithm \ref{algo:algo} to find the best matching from the provided transcription ($T$) for each utterance ($A_k$). In the algorithm, $DIS$ is the Levenshtein distance between two sentences. After this module, two datasets will be generated. The first dataset $\text{DATA}_{align}$ is what the algorithm finds close alignment between the prediction and provided transcription that is within a threshold. 
The second dataset $\text{DATA}_{verify}$ is what the algorithm finds slight mismatches between the prediction and the transcription. For $\text{DATA}_{align}$, the provided transcription from $T$ will be used as the ground truth of the utterance. $\textbf{\textcircled{4}}$ The fourth module, post-generation checking (PGC), is an optional module that iterates through $\text{DATA}_{align}$ to find if there are significant mismatches between a second-round prediction and the aligned transcription on sentence length. The implemented metric for PGC is based on the difference in sentence length between the results of a second-round prediction and the aligned transcription. If the difference is greater than a threshold, the utterance and its transcription will be removed from $\text{DATA}_{align}$. $\textbf{\textcircled{5}}$ The fifth module, user selection, is an optional module that launches a graphical-user-interface (GUI) that allows the user to listen to, select, or input correct transcription for each utterance in $\text{DATA}_{verify}$ so that they could be added to the dataset. After the two optional modules, \mname~ assumes the validity of $\text{DATA}_{align}$, which will be used as the final output dataset.

Currently, speaker identification is not supported by \mname. We choose not to include diarization in \mname~ to ensure the correctness of the alignment. However, if the user has specific needs for diarization, they need to modify the second and third modules to incorporate diarization features. We believe that the modification is relatively easy and straightforward.

\section{CHILES as an Example}

\subsection{CHILDES \cite{CHILDES} dataset}
The Child Language Data Exchange System (CHILDES) \cite{CHILDES} is a component of the TALKBANK project \footnote{\url{https://talkbank.org/}}. CHILDES is a collection of many existing research works and contains a large amount of audio and transcriptions of children's speech under various conditions. Among all of the audio files, we select four datasets that contain audio spoken by children in English and accompanied by English transcriptions. We use \mname~ to convert them into smaller audio/transcription segments that are compatible with the training requirements of the current DL models. We report the specifications of each dataset in Table \ref{table:chi}. Among the four datasets, Narrative records 352 children from ages 4 to 9. The children are performing the Edmonton Narrative Norms Instrument (ENNI) test \cite{ENNI}. To the best of our knowledge, this generated dataset will be the first at-scale dataset for young children with ASR from clinical recordings. Clinical-Eng includes additional ENNI test utterances, with some of them being duplicates of Narrative. Clinical-Other contains conversations between children and clinicians, and the utterances are much shorter, often at the word level. English-NA contains recordings of 5332 conversations in households when a child is involved. Compared to the other datasets, the Eng-NA dataset has varying audio qualities and contains not only children's speech but also adult speech as the audios are recorded in the home setting. During dataset generation, we set $\sigma_a=0.1$, $\sigma_i=0.3$; we also enable post-generation checking with maximum length tolerance as one word. For the backbone model, we use WhisperX \cite{bain2022whisperx} because of its superior performance over other ASR models, but we will also show the performance of another variant \cite{Stable_ts} in the evaluations.

\begin{table}[!ht]
    \centering
    \caption{Specifications for the four datasets extracted from CHILDES \cite{CHILDES}. Original is the number of participants (or the number of audio files in the original dataset). AU is the number of aligned utterances, whereas VU is the number of utterances to be verified if the user chooses the manual selection. Time is the total duration of all the utterances in the aligned dataset, the format of Hour: Minute: Second.}
    \label{table:chi}
    \begin{adjustbox}{width=0.45\textwidth}
    \small
    \begin{tabular}{|l|l|l|l|l|}
    \hline
         Dataset &  Original & AU & VU & Time (H:M:S) \\ \hline
        Narrative & 352 & 14654 & 4402 &15:16:51\\ \hline
        Clinical-Eng & 1540 & 59539 & 14610& 30:11:40\\ \hline
        Clinical-Other & 292 & 21982 & 677 & 4:18:6\\ \hline
        English-NA & 5332 & 778978 & 180146 &285:18:22\\ \hline
    \end{tabular}
    \end{adjustbox}
\end{table}
\vspace{-5mm}
The processed datasets are organized in a similar way as LibriSpeech \cite{librispeech}. Each (anonymized) participant's audio file is segmented, and the segmented audio utterances are put under a folder named by the hashed participant. A transcription is paired with each utterance's audio file. \footnote{Due to restrictions on commercial usage from the raw data, we will not publish the processed dataset. Contact the authors for research usage, or use the software to force-align by yourself.} Currently, the audio utterance is in MP3 format, and the transcription is in TXT format.

\begin{table}[!t]
\centering

    \caption{Manual inspections by the authors on the generation quality of \mname~ on two randomly selected audio files and their transcriptions. AU is the number of aligned utterances in $Data_{align}$, and VU is the number of utterances in $Data_{verify}$. AU Error is the number of utterances that have any incorrect transcription. AW is the number of aligned words, and AW Error is the number of words that are incorrect in the aligned words. The percentage by the AW Error is the WER percentage of the aligned words. PGCU is the number of utterances that are removed from $Data_{align}$ in post-generation checking, and FP in the parentheses is the number of false positives in PGCU.}
    \label{table:eval}
\begin{adjustbox}{width=0.47\textwidth}

\begin{tabular}{|l|l|l|l|l|l|l|}
\hline
Backbone Model          & AU & VU & AU Error  & AW  & AW Error (\%) & PGCU (FP) \\ \hline
Stable whisper \cite{Stable_ts} & 77 & 32 & 2   & 814 & 3   (0.37\%) & 18 (11)    \\ \hline
Whisperx \cite{bain2022whisperx}      & 81 & 33 & 1   & 903 & 2  (\textbf{0.22\%}) & 5 (3)    \\ \hline
\end{tabular}
\end{adjustbox}
\vspace{-5mm}
\end{table}
\subsection{Evaluations}
Due to the size of the dataset, we are not able to perform a manual quality verification on the entire generated dataset. Thus, we randomly selected two audio files and transcriptions from the 352 recordings in the Narrative dataset, and we report the manual inspection results for data generated by \mname~ with these two audio files in Table \ref{table:eval}. Several results are worthy of emphasizing here. First, since the two transcriptions are noisy, MFA \cite{mcauliffe17_interspeech} completely fails to properly align the audio segments with the correct transcription. To be specific, both documents have missing transcriptions corresponding to the beginning of the audio, which results in 100\% AU Error with MFA's ``segment" command and the inability to execute with the ``align" command. Similarly, the alignment function in Batchalign \cite{Liu_MacWhinney_Fromm_Lanzi_2023} assumes the validity of the provided transcription (which in their pipeline would have been generated by a DL model), and produces incorrect alignments. Second, WhisperX \cite{bain2022whisperx} shows better performance than Stable-whisper \cite{Stable_ts}, and given its faster speed, we recommend user to use WhisperX \cite{bain2022whisperx} wherever possible. Third, \mname~ using WhisperX \cite{bain2022whisperx} as the backbone incorrectly aligns one utterance with its transcription. For that utterance, it misses the ``so the'' sound at the end of the utterance, and the two words are not recorded into the aligned transcription. Manual inspection finds that the speaker stuttered and repeated ``so the", which might be the issue of the model not picking up that trailing sound in the segmented utterance. At last, WhisperX \cite{bain2022whisperx} has a WER of \textbf{0.22\%} and Stable-whisper \cite{Stable_ts} has a WER of 0.37\% 
for the aligned words. This result is much better than human annotators. Works from Attia et. al. 
\cite{attia2023kidwhisper} reported that 5 out of 393 hours of speech in MyST dataset \cite{pradhan2023science} are potentially incorrect with WER$>50\%$, resulting in \textbf{3\%} increase in WER for the entire training dataset. Compared to human annotators that were used to annotate MyST, \mname~ achieves one magnitude lower WER (\textbf{13.6$\times$}) without requiring any human labor.

\section{Conclusion}
In this paper, we present the Flexible and Automatic Speech Aligner (FASA) toolkit, which leverages DL models as its backbone to automate the alignment of children's speech datasets with minimal human intervention. At the same time, we introduce the first young children ASR dataset sourced from clinical data, filling a significant gap in available resources. Compared to human annotators in another dataset, alignments produced by FASA achieve \textbf{13.6$\times$} lower WER without human intervention. 
The availability of our toolkit, code, and dataset aims to propel further research and collaboration in advancing ASR for children, ultimately enhancing technology's ability to understand and interact with the unique characteristics of children's speech.

\newpage

\bibliographystyle{IEEEtran}
\bibliography{mybib}

\end{document}